\pdfoutput=1

\documentclass[11pt]{article}

\usepackage{acl}

\usepackage{times}
\usepackage{latexsym}
\usepackage{amsmath,amssymb,amsfonts}
\usepackage{algorithm, algcompatible}
\usepackage{graphicx}
\usepackage{comment}
\usepackage{multirow}
\usepackage{hhline}

\usepackage[T1]{fontenc}

\usepackage[utf8]{inputenc}

\usepackage{microtype}

\usepackage{inconsolata}

\title{N-gram Prediction and Word Difference Representations for Language Modeling}

\author{DongNyeong Heo \\
  Handong Global University\\
  South Korea \\
  \texttt{sjglsks@gmail.com} \\\And
  Daniela Noemi Rim \\
  Handong Global University\\
  South Korea \\
  \texttt{rim.dan96@gmail.com} \\\And
  Heeyoul Choi \\
  Handong Global University\\
  South Korea \\
  \texttt{hchoi@handong.edu} \\}

\begin{document}
\maketitle
\begin{abstract}
Causal language modeling (CLM) serves as the foundational framework underpinning remarkable successes of recent large language models (LLMs). Despite its success, the training approach for next word prediction poses a potential risk of causing the model to overly focus on local dependencies within a sentence. While prior studies have been introduced to predict future $N$ words simultaneously, they were primarily applied to tasks such as masked language modeling (MLM) and neural machine translation (NMT). In this study, we introduce a simple $N$-gram prediction framework for the CLM task. Moreover, we introduce word difference representation (WDR) as a surrogate and contextualized target representation during model training on the basis of $N$-gram prediction framework. To further enhance the quality of next word prediction, we propose an ensemble method that incorporates the future $N$ words' prediction results. Empirical evaluations across multiple benchmark datasets encompassing CLM and NMT tasks demonstrate the significant advantages of our proposed methods over the conventional CLM. 
\end{abstract}

\section{Introduction}
With the remarkable advancements in deep learning techniques, neural language modeling has become a central component in modern natural language processing (NLP) tasks, such as natural language understanding (NLU), neural machine translation (NMT) and question answering. Among the approaches to language modeling, causal language modeling (CLM), which predicts the next word given the previous words, is a widely employed language modeling framework. For example, prominent large language models (LLMs) like GPT-2 \citep{radford2019language} and GPT-3 \citep{brown2020language} rely on CLM as their primary training framework. Despite their successful applications, the prevalent next word prediction manner can inadverently lead models to overfit to local dependencies rather than capturing long-term dependencies between words. This tendency arises from some phrases or paired words that have strong dependencies with each other, such as "\textit{Barack Obama}" and "\textit{Harry Potter}" \citep{qi2020prophetnet}. 

A way of mitigating this problem involves predicting not solely the next word but also subsequent words in later time-steps such as $N$-gram prediction. Researchers \citep{sun2019ernie, joshi2020spanbert, xiao2020ernie, qi2020prophetnet} have adopted this $N$-gram prediction methodology for the masked language modeling (MLM) during the pre-training phase of LLMs \cite{devlin2018bert}. Similar approaches have been applied to the NMT task \citep{shao2018greedy, ma2018bag, shao2020minimizing}. However, these methods often require significant modifications to the model architecture, a different loss function than the conventional cross-entropy loss, or an expansion of the vocabulary for $N$-grams.

\begin{figure*}[t]
    \centering 
    \includegraphics[width=0.88\linewidth]{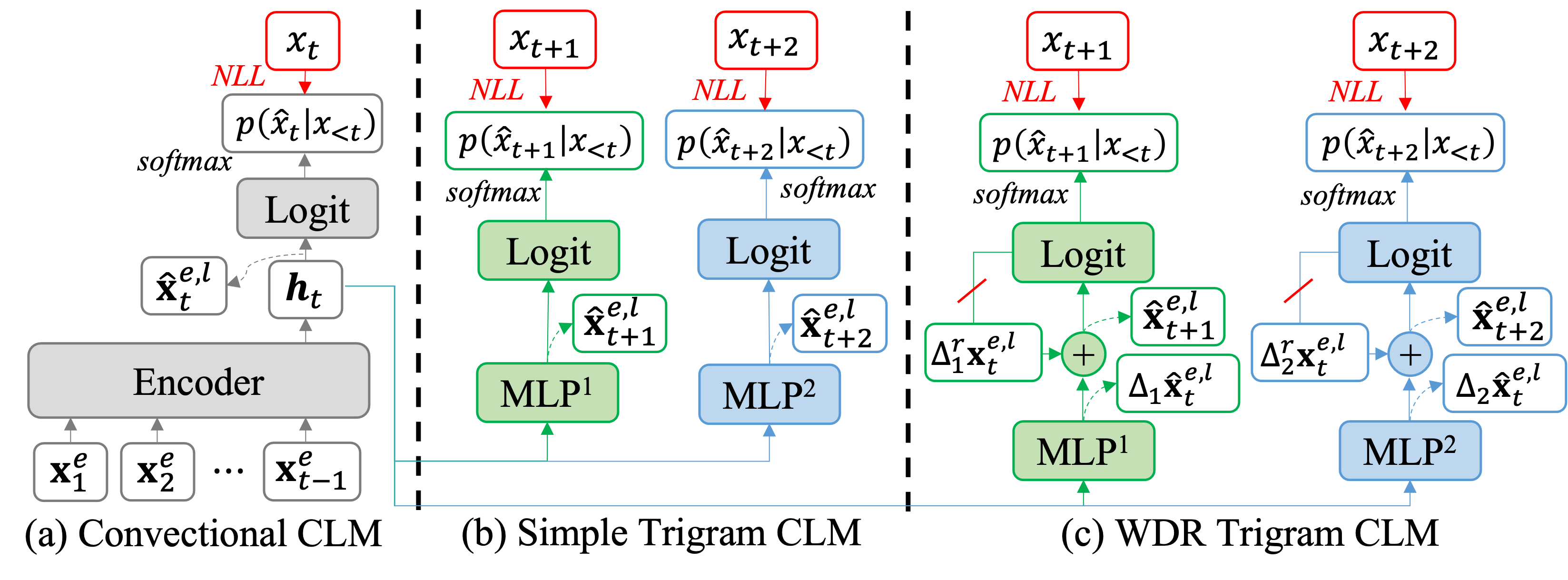}
    \caption{Model illustrations of (a) conventional CLM, (b) simple $N$-gram CLM, and (c) WDR $N$-gram CLM when $N=3$. Note that all of the drawn logit layers above the MLP layers are the same function with the same parameter. Red diagonal lines in (c) on lines from logit layer to $\Delta^r_i\mathbf{x}^{e,l}_t$ indicate detaching operation.}
    \label{fig:model_illustration}
\end{figure*} 

This paper introduces a novel $N$-gram prediction framework designed specifically for CLM and proposes innovative methods aimed at fortifying this framework. The contributions of this work can be summarized as follows.
\textit{(1) A simple $N$-gram prediction for CLM}: we propose a simple $N$-gram prediction integrated to existing CLM models. Except for an additional multi-layer perceptron (MLP) layer, our method does not require other modifications to model architecture, loss function, and vocabulary.
\textit{(2) Word difference representation}: we propose to use the embedding vectors' difference between contiguous words, termed word difference representation (WDR), as a surrogate representation for individual words. Departing from the conventional approaches that employing a fixed word embedding as target representation, we provide diverse WDR as target representations in accordance with context. We discovered this method can vary backpropagated gradient during training so that it can enhance generalizability. The algorithmic reversibility of WDR preserves the feasibility of the above simple $N$-gram prediction method.
\textit{(3) An ensemble method suitable for the CLM task}: we propose an ensemble method designed to refine the next word prediction by leveraging other multiple $N$ predictions from the $N$-gram prediction.

Our preliminary and primary experimental results, conducted several CLM benchmark datasets, highlight the gradual improvements in perplexity achieved by our proposed simple $N$-gram framework, the WDR diverse target representations, and ensemble method when compared to several baseline models. Our qualitative analysis focusing on gradient elucidates the advantage of the WDR method from the perspective of optimization generalizability. In addition to the main CLM task, we demonstrate the applicability and advantages of our proposed approaches to the NMT task, which is a conditional form of the CLM task.

\section{Background: Conventional CLM}
\label{sec:conventional_clm}
Since the work of \citep{bengio2000neural}, neural network-based language modeling has been developed and become mainstream in language modeling. As background knowledge, we describe the conventional training framework of CLM (the next word prediction) in this section.

A sentence consists with words, $X=\{x_1, x_2, \cdots, x_{T}\}, x \in \mathcal{V}$, where $T$ means the sequence length of the sentence and $\mathcal{V}$ is the vocabulary set. Conventional CLM computes the likelihood of a word conditioned on its preceding words in the sentence, $p(x_t|x_{<t})$. For processing, words are mapped to embedding vectors \citep{mikolov2013efficient}, and the encoded hidden state at time-step $t$ is formulated as follows:
\begin{equation}\label{eq:clm_encoder}
    \mathbf{h}_t=Enc_{\theta}(\{\mathbf{x}^e_1,\mathbf{x}^e_2, \cdots, \mathbf{x}^e_{t-1}\}) \in \mathbb{R}^{d},
\end{equation}
where $\mathbf{x}^e_t \in \mathbb{R}^{d}$ means the embedded vector of $x_t$. $Enc_{\theta}$ is an encoder model with its parameter set $\theta$. $d$ is the dimension of the encoded hidden state and the embedding vector spaces. Recently, most language models use Transformer \citep{vaswani2017attention} as their encoder architecture. After encoding, the encoded hidden state is linearly transformed to a logit value of each word in a vocabulary set $\mathcal{V}$. Finally, the likelihood of the predicted word is formulated as follows:
\begin{align}
    p(\hat{x}_t|x_{<t};\theta) &= softmax(\hat{\mathbf{x}}^{l}_t), \nonumber \\
    \hat{\mathbf{x}}^{l}_t &= \mathbf{W}^l\mathbf{h}_t=\mathbf{W}^l\hat{\mathbf{x}}^{e,l}_t, \label{eq:predicted_logit_emb}
\end{align}
where $\mathbf{W}^l \in \mathbb{R}^{|\mathcal{V}| \times d}$ is the weight matrix of the logit layer. 

To help the understanding of our idea, we note that a parameter vector of logit layer's weight is another word embedding set that is mapped to the target word, that is $\mathbf{W}^l=[\mathbf{x}^{e,l}_1,\mathbf{x}^{e,l}_2,\cdots,\mathbf{x}^{e,l}_{|\mathcal{V}|}]^\top$. In this point of view, the encoded hidden state, $\mathbf{h}_t$, is the predicted word embedding vector of the logit layer, $\hat{\mathbf{x}}^{e,l}_t$. Then, the inner product between $\mathbf{W}^l$ and $\hat{\mathbf{x}}^{e,l}_t$ outputs the predicted score of each embedding that indicates how the predicted word embedding is similar to the logit layer's original word embedding.

Finally, the model learns to minimize the negative log-likelihood (NLL) loss as follows:
\begin{equation}\label{eq:negative_log_likelihood}
    \mathcal{L}(X,\theta)=-\sum_{t=1}^T\log p(\hat{x}_t=x_t|x_{<t};\theta).
\end{equation}
This loss becomes the minimum when the model exactly predicts the logit layer's embedding of the target word, that is $\hat{\mathbf{x}}^{e,l}_t=\mathbf{x}^{e,l}_t$. This process is illustrated in Fig.\ref{fig:model_illustration}(a).

\section{Proposed Methods}
In this section, we propose three ideas: (1) a simple $N$-gram CLM, (2) word difference representation $N$-gram CLM, and (3) an ensemble method over $N$-gram predictions. 

\subsection{Simple $N$-gram CLM}
\label{subsec:simple_n_gram_clm}
First, we propose a simple $N$-gram prediction on the conventional framework of CLM. The core idea is adding an MLP layer to predict a future word given the same hidden state of the conventional CLM. This process is formulated as follows:
\begin{equation}\label{eq:n_th_mlp}
    \hat{\mathbf{x}}^{e,l}_{t+n}=MLP^n(\mathbf{h}_t).
\end{equation}
For instance, assuming $N$ is 3, two MLP layers, $MLP^1$ and $MLP^2$, are employed and predict $\hat{\mathbf{x}}^{e,l}_{t+1}$ and $\hat{\mathbf{x}}^{e,l}_{t+2}$, respectively, as shown in Fig.\ref{fig:model_illustration}(b).
The limited capability of the MLP layer to learn an effective function from a large and complicated dataset may regularize the main encoder, $Enc_{\theta}$, to encode a simultaneously informative hidden state for all $N$-gram predictions. This regularization might be beneficial to prevent the model to overly focus on local dependencies.

We compute the likelihoods of the future target words, $p(\hat{x}_{t+1}|x_{<t};\theta)$ and $p(\hat{x}_{t+2}|x_{<t};\theta)$ in the above example, following each logit layer and the softmax function. Instead of using individual logit layers for each future word prediction, we share the parameters of all logit layers, including the conventional CLM model's logit layer. Therefore, this approach increases just a small amount of parameters for each additional MLP layer. Furthermore, it re-uses the original (unigram) vocabulary set for the future word prediction, not an additional large vocabulary set of $N$-grams.
The loss for $n$-th future word prediction is as follows:
\begin{equation}\label{eq:future_n_nll}
    \mathcal{L}_{n}(X,\theta)=-\sum^{T-n}_{t=1}\log p(\hat{x}_{t+n}= x_{t+n}|x_{<t};\theta).
\end{equation}
As like Eq.\eqref{eq:negative_log_likelihood}, this loss becomes minimum when the model exactly predicts the future target word's embedding, i.e., $\hat{\mathbf{x}}^{e,l}_{t+n}=\mathbf{x}^{e,l}_{t+n}$. The total loss for the training of this simple $N$-gram CLM model is the mixture of Eq.\eqref{eq:negative_log_likelihood} and Eq.\eqref{eq:future_n_nll} as follows:
\begin{equation}\label{eq:total_loss}
    \mathcal{L}^{tot}_{N}(X,\theta)=\frac{1}{2}\mathcal{L}(X,\theta)+\frac{1}{2(N-1)}\sum^{N-1}_{i=1}\mathcal{L}_{i}(X,\theta).
\end{equation}
Notably, we do not equally take the average of the original loss, Eq.\eqref{eq:negative_log_likelihood}, with other losses, since the next word typically has stronger dependencies with the preceding words than other future words. In other words, averaging the entire set of loss terms together might introduce excessive regularization.

\subsection{Word Difference Representation (WDR) $N$-gram CLM}
To use a more informative target than simple $N$-gram CLM, we introduce the idea of WDR which is a contextualized surrogate representation of words within a sentence. Basically, it is based on a form of word embedding compositions: the difference vector, $\mathbf{x}^e_{t+1}-\mathbf{x}^e_t$. Since \citep{mikolov2013efficient} demonstrated that arithmetic compositions of learned word embedding can convey semantic meanings, many researches have explored the word embedding compositionality \citep{xu2015word, hartung2017learning, poliak2017efficient, scheepers2018improving, li2018subword, frandsen2019understanding}. Their studies utilized composed word embeddings as inputs to models, instead of original word embeddings, showcasing their advantages across various NLP tasks.

Unlike the prior research, we provide WDR to the model as the target to predict, rather than utilizing it as input. The difference vector of contiguous words offers a different representation for the word depending on its adjacent words. Therefore, by leveraging WDR as the target, we expect the model can learn more diverse targets than previous works. Generating WDR is simple repetition of vector subtractions which is computationally cheap and easy to parallelize, so it does not impose a high computational cost. Moreover, generating WDR is reversible, so that original embedding vectors can be reconstructed from WDR. This property facilitates the development of WDR-based $N$-gram CLM integrating the same framework of the simple $N$-gram CLM without a significant modification. Detailed explanations elucidating these advantages are provided in the subsequent sections.

\subsubsection{Definition of $n$-level WDR}
\label{subsubsec:definition_wdr}
As we briefly mentioned above, we use the difference of contiguous embedding vectors as the base of WDR. Given an embedding vector sequence $\{\mathbf{x}^e_1,\mathbf{x}^e_2,\cdots,\mathbf{x}^e_T\}$, the 1-level WDR at the time-step $t$ is defined as follows:
\begin{align}
    \Delta_1\mathbf{x}^e_t&=
        \begin{cases}
            \mathbf{x}^e_{t+1}-\mathbf{x}^e_t & \text{if } 1\le t < T, \\
            \mathbf{x}^e_T & \text{if } t = T. \\
        \end{cases} \label{eq:1level_wdr}
\end{align}
In an inductive manner, the $n$-level WDR at the time-step $t$ when $n>1$ is defined as follows:
\begin{align}
    \Delta_{n}\mathbf{x}^e_t&=
        \begin{cases}
            \Delta_{n-1}\mathbf{x}^e_{t+1}-\Delta_{n-1}\mathbf{x}^e_t & \text{if } 1\le t < T, \\
            \Delta_{n-1}\mathbf{x}^e_T=\mathbf{x}^e_T & \text{if } t = T. \\
        \end{cases} \label{eq:n_level_wdr}
\end{align}

As an alternative of the above $n$-level WDR definition, we explored the opposite direction to subtract the contiguous vectors, that is $\Delta_{n-1}\mathbf{x}^e_{t}-\Delta_{n-1}\mathbf{x}^e_{t+1}$. In our internal empirical studies, we discovered the alternative design achieved similar performances. Therefore, we follow the design of Eq. \ref{eq:n_level_wdr} throughout this paper.

Based on the definitions of Eqs. \ref{eq:1level_wdr} and \ref{eq:n_level_wdr}, the $n$-level WDR can be represented by the composition of original word embeddings. For example, the 2 and 3-level WDRs at time-step $t$ can be represented as follows: $\Delta_2\mathbf{x}^e_t=\mathbf{x}^e_{t+2}-2\mathbf{x}^e_{t+1}+\mathbf{x}^e_t$ and $\Delta_3\mathbf{x}^e_t=\mathbf{x}^e_{t+3}-3\mathbf{x}^e_{t+2}+3\mathbf{x}^e_{t+1}-\mathbf{x}^e_t$, respectively. With this manner, we can derive the formulation of $n$-level WDR as follows:
\begin{equation}\label{eq:n_level_WDR_w_orgemb}
    \Delta_n\mathbf{x}^e_t=\sum^{n}_{i=0}\binom{n}{i}(-1)^i\mathbf{x}^e_{t+(n-i)},
\end{equation}
where $\binom{n}{i}=\frac{n!}{(n-i)!i!}$ is the binomial coefficient. This equation holds for every positive integer of $n$ 
and for every time-step $t$ when $ t \leq T-n$.
See Appendix \ref{subsec:proof_n_level_wdr_w_orgemb} for a proof of this equation.

As we mentioned earlier, $n$-level WDR is reversible to the original word embedding. For the 1-level WDR, $\mathbf{x}^e_{t+1}$ can be reconstructed by adding $\mathbf{x}^e_t$ to $\Delta_1\mathbf{x}^e_t$. Likewise, $\mathbf{x}^e_{t+n}$ can be reconstructed by adding $-\sum^{n}_{i=1}\binom{n}{i}(-1)^i\mathbf{x}^e_{t+(n-i)}$ to $\Delta_n\mathbf{x}^e_{t}$ (note that the first term of the right-hand side of Eq.\eqref{eq:n_level_WDR_w_orgemb} is $\mathbf{x}^e_{t+n}$). For simplicity, we use a new notation for the conjugate term that reconstructs the original embedding by addition to the $n$-level WDR as follows:
\begin{equation}\label{eq:reconstruction_term}
    \Delta^r_{n}\mathbf{x}^e_t=-\sum^{n}_{i=1}\binom{n}{i}(-1)^i\mathbf{x}^e_{t+(n-i)},
\end{equation}
This leads to $\Delta_{n}\mathbf{x}^e_{t}+\Delta^r_{n}\mathbf{x}^e_{t}=\mathbf{x}^e_{t+n}$. The conjugate term for reconstruction, $\Delta^r_{n}\mathbf{x}^e_{t}$, can be obtained by Eq.\eqref{eq:reconstruction_term} or  iterative operations of Eq.\eqref{eq:n_level_wdr}.

\subsubsection{Training of WDR $N$-gram CLM}
\label{subsubsec:wdr_ngram_clm}
We develop the WDR-based $N$-gram CLM from the framework of simple $N$-gram CLM.
To achieve the mentioned goal that providing the WDR as the target of the model, we apply the definitions and derivations in Sec.\ref{subsubsec:definition_wdr} to the logit layer's embeddings.
Following the idea of the simple $N$-gram CLM described in Sec.\ref{subsec:simple_n_gram_clm}, we employ MLP layers for predictions of $N$-gram. However, in WDR $N$-gram CLM, the $MLP^n$ layer outputs $\Delta_n\hat{\mathbf{x}}^{e,l}_{t}$ instead of $\hat{\mathbf{x}}^{e,l}_{t+n}$. Then we produce its corresponding conjugate term, $\Delta^r_{n}\mathbf{x}^{e,l}_{t}$, based on the logit layer's embedding matrix. Adding those two, $\Delta_{n}\hat{\mathbf{x}}^{e,l}_{t}+\Delta^r_{n}\mathbf{x}^{e,l}_{t}$, yields $\hat{\mathbf{x}}^{e,l}_{t+n}$ as in the simple $N$-gram CLM. Then, we take the same processes of the logit, likelihood, and loss computations as in the simple $N$-gram CLM.

An essential design of this framework is detachment of the produced conjugate term, $\Delta^r_{n}\mathbf{x}^{e,l}_{t}$, from the backpropagation process. Absence of this detachment might lead the model to adjust the logit layer's weight matrix in a distorted manner, because the input of the logit layer is recursively produced from itself. 

In WDR $N$-gram CLM, the minimum value of NLL loss of $x_{t+n}$ prediction, Eq.\eqref{eq:future_n_nll}, is achieved when $\hat{\mathbf{x}}^{e,l}_{t+n}=\mathbf{x}^{e,l}_{t+n}$, which is $\Delta_{n}\hat{\mathbf{x}}^{e,l}_{t}+\Delta^r_{n}\mathbf{x}^{e,l}_{t}=\Delta_{n}\mathbf{x}^{e,l}_{t}+\Delta^r_{n}\mathbf{x}^{e,l}_{t}$ based on the equation led by Eq.\eqref{eq:reconstruction_term}. Because the conjugate term, $\Delta^r_{n}\mathbf{x}^{e,l}_{t}$, is detached, the model would learn to predict $\Delta_{n}\mathbf{x}^{e,l}_{t}$, which is true $n$-level WDR. 
In other words, WDR $N$-gram CLM learns to predict composed word embeddings, offering diverse and contextualized target representations, even for the same target word. The entire process of WDR trigram CLM example is illustrated in Fig.\ref{fig:model_illustration}(c). 

\begin{figure*}[t]
    \centering 
    \includegraphics[width=1.0\linewidth]{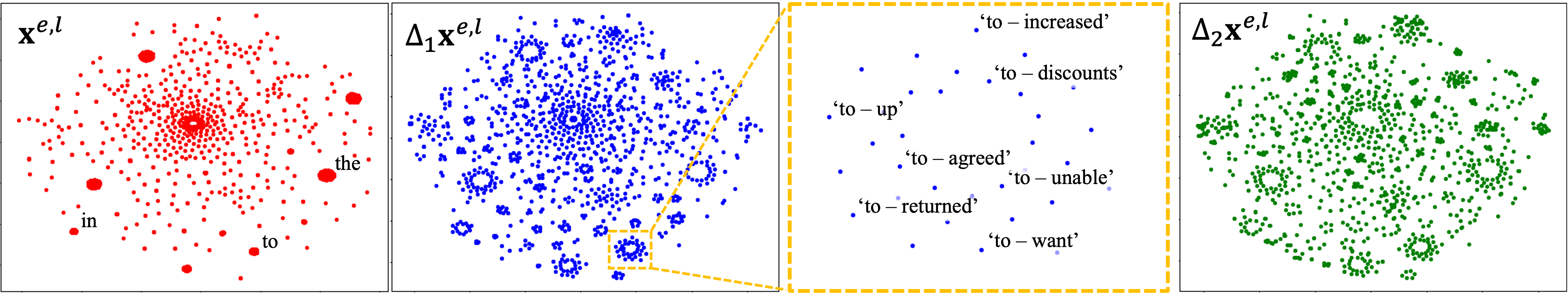}
    \caption{From the left-to-right, they are visualizations of the original embeddings (first), 1-level WDR and the plot zoomed in around the original word `to' (second and third), and 2-level WDR (last), respectively. In the third plot, (`to-\textit{word}') means the 1-level WDR vector, that is $\mathbf{x}^{e,l}_{to}-\mathbf{x}^{e,l}_{word}$ based on the word `\textit{to}' fragment within the sentence.}
    \label{fig:wdr_visualizations}
\end{figure*} 

\subsubsection{How Diverse Are WDR-based Target Representations?}
\label{subsubsec:visualizations_of_wdr}
In order to gain a more profound understanding of WDR as target representations, we explored how WDR would diversify target representations compared to the conventional CLM or the simple $N$-gram CLM. As we mentioned in Sec.\ref{sec:conventional_clm} and Sec.\ref{subsec:simple_n_gram_clm}, the conventional CLM and the simple $N$-gram CLM utilize the logit layer's embeddings as target representations to predict. To see the practical examples of these target representations, we collected 1,270 representations from the logit layer's embedding matrix of the pre-trained conventional CLM model (`TF' in the preliminary experiment, Sec.\ref{subsubsec:preliminary_analysis}). The 1,270 representations correspond to all the tokens of randomly selected 10 sentences from the Penn TreeBank (PTB) \citep{mikolov2014learning} testset. Also, we computed 1 and 2-level WDRs with the collected embeddings, and added them to the collection, resulting in 3,810 representations in total. Finally, we reduced the dimension of the total collection to 2-dimension with t-SNE algorithm \citep{van2008visualizing}. 

Fig.\ref{fig:wdr_visualizations} shows the collected representations in a 2-dimensional space. The first plot illustrates the original embeddings, $\mathbf{x}^{e,l}$. Note that the representations of frequent words, such as `to' may be included more times than other words in the collection. We interpret that this is the reason why t-SNE places frequent words (e.g., `in', `to', and `the') distant from other less frequent words to resemble the non-uniform distribution of the collection. 
On the other hand, the 1-level WDR representations, $\Delta_1\mathbf{x}^{e,l}$, look more diverse compared to the original embeddings as in the second plot. For example, by composing adjacent words such as `want', `unable', `returned', into the frequent word `to', it diversifies the embedding representations according to its previous word as in the third plot which is zoomed in. The 2-level WDR looks more diverse even compared to 1-level WDR as in the last plot. Based on this analysis, we expect WDR $N$-gram CLM to give more diverse target representations than other methods, such as conventional CLM and the simple $N$-gram CLM.

\subsection{Ensemble Method to Refine the Next Word Prediction Leveraging $N$-gram Predictions}
\label{subsec:ensemble_method}
We propose a new ensemble method to incorporate the $N$-gram predictions into the process of the next word prediction. The encoder model, such as Transformer, outputs $\{\mathbf{h}_2,\mathbf{h}_3,\cdots,\mathbf{h}_t\}$ given the embedded input sentence $\{\mathbf{x}^e_1,\mathbf{x}^e_2,\cdots,\mathbf{x}^e_{t-1}\}$. The encoded hidden state $\mathbf{h}_i$ represents the computed hidden state given the inputs up to time-steps $(i-1)$. At testing, in addition to the predicted embedding $\hat{\mathbf{x}}^{e,l}_t$ from the conventional CLM, $MLP^n$ layer of $N$-gram CLM can estimate the target word for time $t$ given $\mathbf{h}_{t-n}$. Therefore, we can get $N$ predicted embeddings for the current time-step. We ensemble these predicted embeddings just before the logit layer using the following formulation:
\begin{equation}\label{eq:ensemble_emb}
    \hat{\mathbf{x}}^{e,l}_{t,ens}=(1-\lambda)\hat{\mathbf{x}}^{e,l}_t+\frac{\lambda}{N-1}\sum^{N-1}_{i=1}MLP^i(\mathbf{h}_{t-i}),
\end{equation}
where
$\lambda$ is a scalar value between 0 and 1. It controls the influences of future word predictions (but derived from past time-steps) on the current word prediction. Similar to the rationale behind the dominance of the original NLL loss in its total loss formulation, Eq.\eqref{eq:total_loss}, we do not equally average the original predicted embedding with others. In the case of WDR-based $N$-gram CLM, we ensemble $MLP^i(\mathbf{h}_{t-i})+\Delta^{r}_{i}\mathbf{x}^{e,l}_{t-i}=\hat{\mathbf{x}}^{e,l}_{t}$ in the summation part in Eq.\eqref{eq:ensemble_emb}.

After this ensemble computation, we input it to the logit layer and compute the next word's likelihood. At testing, this ensemble likelihood result is used to compute perplexity (PPL) in CLM tasks or serving as candidate scores for beam search in NMT tasks.

\section{Experiments and Results}
To assess the performances of our proposed methods, we conducted CLM and NMT experiments on multiple benchmark datasets. 

\subsection{Causal Language Modeling (CLM)}
For the CLM task, we executed two experiments: preliminary and primary. The preliminary experiment was dedicated to monitor the dynamics of two hyperparameters: $N$ and $\lambda$ toward the performance. In contrast, we only report the results of the best hyperparameters in the primary experiment's demonstration.

\label{subsec:clm_experimental_results}
\subsubsection{Data Description}
PTB (-, 0.9M tokens, 10K vocabulary), WikiText-2 (W2, 2M tokens, 33K vocabulary), Text8 (T8, 15M tokens, 254K vocabulary), and WikiText-103 (W103, 103M tokens, 268K vocabulary) \citep{mikolov2014learning, merity2016pointer}. To ensure standardization and transparency in our data-related processes (e.g., download, tokenization, vocabulary, and train/valid/testsets splitting), we relied on open sources. Specifically, the W2 and T8 datasets were sourced from the GitHub repository\footnote{https://github.com/chakki-works/chazutsu}, while the PTB and W103 datasets were sourced from the Tensorized Transformer \cite{ma2019tensorized}'s GitHub repository\footnote{https://github.com/szhangtju/The-compression-of-Transformer}. In the primary experiment, we used the whole datasets, whereas the preliminary experiment was conducted solely on the PTB dataset.

\subsubsection{Models and Training}
\label{subsubsec:clm_models}
For the baseline model of the preliminary experiment, we implemented Transformer (TF) encoder-based CLM. The total number of parameters of the TF baseline is 12M, and our proposed simple and WDR methods increase only 0.1M parameters per an additional MLP layer (note that the logit layer's parameters are all shared). The details of model architecture and training method for the preliminary experiment are described in Table~\ref{table:setting_for_transformers} (in Appendix \ref{subsec:experimental_settings}) in the column of `Small Enc. TF CLM'.

For the baseline models of the primary experiment, we trained the two baseline models that are advanced ones based on TF: tensorized transformer (TT) \citep{ma2019tensorized} and Reformer (RF)\footnote{https://github.com/lucidrains/reformer-pytorch} \cite{kitaev2020reformer}. We mostly followed their reported configurations, except some minor changes such as the number of tokens in a mini-batch and learning rates. The details of these changes for each dataset are described in Table~\ref{table:changed_settings_for_primary_baselines} (in Appendix \ref{subsec:experimental_settings}). As a result, the total numbers of parameters of (TT, RF) models according to datasets are (6.7M, 15.3M) for PTB and W2, (82.4M, 236.6M) for T8 and W103, respectively. Our proposed simple and WDR methods increase the number of parameters by 0.1M and 0.5M, respectively, per an additional MLP layer regardless of the type of dataset.

On top of the baseline models, we applied our proposed method, and we call them `TF+Sim', `TF+WDR', `TT+Sim', `TT+WDR', `RF+Sim', and `RF+WDR'. We varied $N$ from 2 to 4 and $\lambda$ from 0.0 to 0.6 for every experiment of our proposed methods. In the demonstration of the primary experiment results, we report the result of the best hyperparameter setting of each model. These settings are reported in the `CLM Task' column of Table~\ref{table:configurations_of_our_approach} (in Appendix \ref{subsec:experimental_settings}).

\begin{table}[]
\centering
\caption{Word-level PPL results of the preliminary experiment with Transformer encoder-based CLMs on the PTB dataset. A different value of $\lambda$ indicates the application of the proposed ensemble method with the $\lambda$ value.}
\resizebox{0.45\textwidth}{!}{
\begin{tabular}{|c|cccc|}
\hline
\multirow{2}{*}{Model} & \multicolumn{4}{c|}{Test PPL} \\ 
\cline{2-5}  & \multicolumn{1}{c|}{$\lambda$=0.0}  & \multicolumn{1}{c|}{0.2} & \multicolumn{1}{c|}{0.4} & 0.6 \\ 
\hhline{|=====|} 
TF  & \multicolumn{1}{c|}{161.0} & \multicolumn{1}{c|}{-} & \multicolumn{1}{c|}{-} & - \\ 
\hline
\begin{tabular}[c]{@{}c@{}}TF+Sim $N$=2\\ $N$=3\\ $N$=4\end{tabular}    & \multicolumn{1}{c|}{\begin{tabular}[c]{@{}c@{}}150.8\\ 153.3\\ 158.1\end{tabular}} & \multicolumn{1}{c|}{\begin{tabular}[c]{@{}c@{}}134.6\\ 134.4\\ 133.6\end{tabular}} & \multicolumn{1}{c|}{\begin{tabular}[c]{@{}c@{}}135.3\\ 133.0\\ 129.1\end{tabular}} & \begin{tabular}[c]{@{}c@{}}156.3\\ 151.9\\ 147.1\end{tabular} \\ \hline
\begin{tabular}[c]{@{}c@{}}TF+WDR $N$=2\\ $N$=3\\ $N$=4\end{tabular} & \multicolumn{1}{c|}{\begin{tabular}[c]{@{}c@{}}149.0\\ 153.1\\ 150.5\end{tabular}} & \multicolumn{1}{c|}{\begin{tabular}[c]{@{}c@{}}136.5\\ 136.1\\ 131.6\end{tabular}} & \multicolumn{1}{c|}{\begin{tabular}[c]{@{}c@{}}129.8\\ 128.2\\ 124.1\end{tabular}} & \begin{tabular}[c]{@{}c@{}}\textbf{128.1}\\ 128.8\\ 127.5\end{tabular} \\ \hline
\end{tabular}}
\label{table:preliminary_results}
\end{table}

\subsubsection{Preliminary Experimental Results}
\label{subsubsec:preliminary_analysis}
Table~\ref{table:preliminary_results} presents the outcomes of the preliminary experiments. We trained the model of each configuration five times with different seeds, and we report the average PPL scores. Both `TF+Sim' and `TF+WDR' surpass the performances of the conventional CLM baseline. This observation aligns with findings from previous studies on other tasks \citep{sun2019ernie, joshi2020spanbert, xiao2020ernie, qi2020prophetnet}. The ensemble method consistently improves performance compared to the non-ensemble ones (where $\lambda$=0.0). It usually achieves the best scores at $\lambda$=0.4 for both the `TF+Sim' and `TF+WDR' models. Also, we observed that the `TF+WDR' model maintains strong performance even at $\lambda$=0.6, while the `TF+Sim' model does not. This implies that `TF+WDR' generally generates more accurate predictions for future words. Moreover, `TF+WDR' tends to outperform their `TF+Sim' counterparts in each setting. These findings collectively suggest that the WDR training approach offers benefits over $N$-gram prediction methodologies.

\begin{figure}[]
    \centering 
    \includegraphics[width=0.95\linewidth]{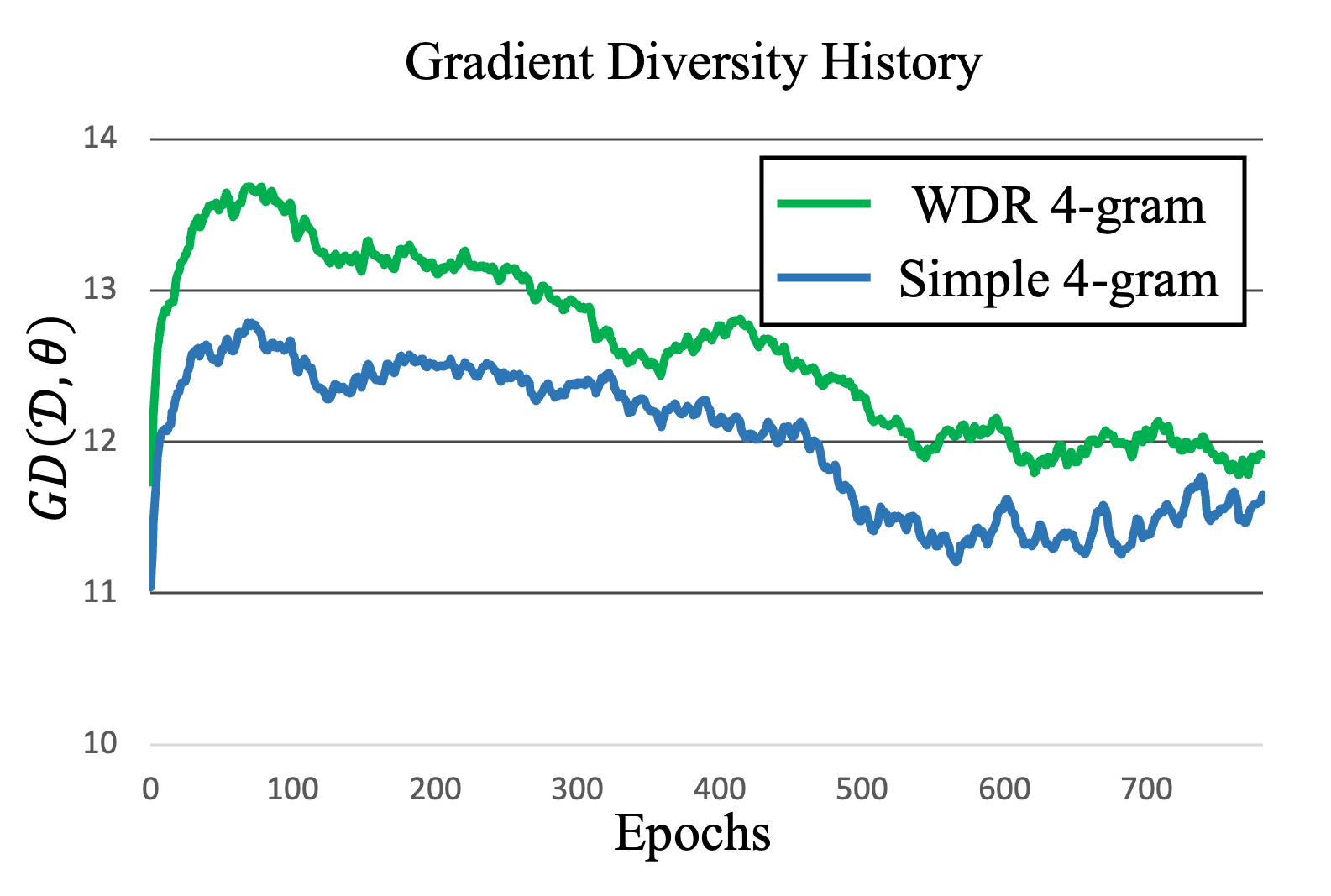}
    \caption{Gradient diversity comparison between simple 4-gram CLM and WDR 4-gram CLM.}
    \label{fig:gradient_diversity}
\end{figure}

\subsubsection{Gradient Diversity Analysis}
As an additional exploration of the advantages of WDR, we checked the connection between the diverse target representations and its benefit during training. Given the evidence in Sec.\ref{subsubsec:visualizations_of_wdr} that WDR gives more diverse target representations compared to other CLMs, it is plausible to guess the backpropagated gradients are also diverse. To quantify this property, we measured `\textit{gradient diversity (GD)}' \citep{yin2018gradient} which is formulated as follows:
\begin{align}
    GD(\mathcal{D},\theta)&=\frac{\sum^{|\mathcal{D}|}_{i=1}||g_i||^2_2}{||\sum^{|\mathcal{D}|}_{i=1}g_i||^2_2},\nonumber \\
    &= \frac{\sum^{|\mathcal{D}|}_{i=1}||g_i||^2_2}{\sum^{|\mathcal{D}|}_{i=1}||g_i||^2_2+\sum_{i\ne j}\langle g_i,g_j \rangle}, \label{eq:gradient_diversity} \\
    g_i&=\nabla_{\theta}\mathcal{L}^{tot}_{N}(X_i,\theta), \nonumber
\end{align}
where $\mathcal{D}=\{X_1,X_2, \cdots,X_{|\mathcal{D}|}\}$ is a mini-batch, $||\cdot||^2_2$ is the squared $L^2$ norm operation, $\langle \cdot,\cdot \rangle$ is the inner product operation, and $\nabla_{\theta}$ is gradient operator with respect to $\theta$. This metric is large when the inner product terms in denominator are small, which means the gradients are different from each other. 

We measured GD of the `TF+Sim $N$=4' and `TF+WDR $N$=4' models in Table~\ref{table:preliminary_results} during training. The GDs over epochs are presented in Fig.\ref{fig:gradient_diversity}. `TF+WDR $N$=4' usually has higher GD than `TF+Sim $N$=4'. As the stochastic property of stochastic gradient descent is known for noisy gradient which enhances generalizability compared to full-batch gradient descent \citep{hardt2016train, yin2018gradient}, higher GD may offer similar advantages due to the stochastic property. Given this understanding, we believe WDR-based training could be beneficial for improving generalization.

\begin{table}[]
\centering
\caption{Word-level PPL results of the primary experiment. Regarding the unsatisfying PPL of `RF (baseline)' on W103, as in the experiments on PTB, W2, and T8 datasets, we trained `RF' on W103 based on the same provided source code with the default configuration except a few changes described in Table \ref{table:changed_settings_for_primary_baselines}. Note that `RF+Sim' and `RF+WDR' models were trained under the same setting for fair comparisons.}
\resizebox{0.4\textwidth}{!}{\begin{tabular}{|c|cccc|}
\hline
\multirow{2}{*}{Model}  & \multicolumn{4}{c|}{Test Word-level PPL} \\ 
\cline{2-5} & \multicolumn{1}{c|}{PTB}  & \multicolumn{1}{c|}{W2}  & \multicolumn{1}{c|}{T8} & W103 \\ 
\hhline{|=====|}
TT (baseline)  & \multicolumn{1}{c|}{55.0}                                                & \multicolumn{1}{c|}{56.1}                                       & \multicolumn{1}{c|}{121.4}                                               & 20.1                                       \\ \hline
\begin{tabular}[c]{@{}c@{}}TT+Sim\\ Ensemble\end{tabular} & \multicolumn{1}{c|}{\begin{tabular}[c]{@{}c@{}}51.6\\ 45.5\end{tabular}} & \multicolumn{1}{c|}{\begin{tabular}[c]{@{}c@{}}62.0\\ 56.0\end{tabular}} &  \multicolumn{1}{c|}{\begin{tabular}[c]{@{}c@{}}106.5\\ \textbf{89.5}\end{tabular}}                                                  & \multicolumn{1}{c|}{\begin{tabular}[c]{@{}c@{}}17.1\\ 17.9\end{tabular}}                                                   \\ \hline
\begin{tabular}[c]{@{}c@{}}TT+WDR\\ Ensemble\end{tabular}       & \multicolumn{1}{c|}{\begin{tabular}[c]{@{}c@{}}47.5\\ \textbf{44.4}\end{tabular}} & \multicolumn{1}{c|}{\begin{tabular}[c]{@{}c@{}}57.7\\ \textbf{53.8}\end{tabular}} & \multicolumn{1}{c|}{\begin{tabular}[c]{@{}c@{}}91.7\\ 90.2\end{tabular}} & \begin{tabular}[c]{@{}c@{}}\textbf{16.8}\\ 16.9\end{tabular} \\ \hline
RF (baseline)                                                               & \multicolumn{1}{c|}{28.0}                                       & \multicolumn{1}{c|}{31.6}                                                & \multicolumn{1}{c|}{64.3}                                                & 50.3                                                \\ \hline
\begin{tabular}[c]{@{}c@{}}RF+Sim\\ Ensemble\end{tabular} & \multicolumn{1}{c|}{\begin{tabular}[c]{@{}c@{}}27.8\\ 26.4\end{tabular}} & \multicolumn{1}{c|}{\begin{tabular}[c]{@{}c@{}}31.6\\ 31.0\end{tabular}} & \multicolumn{1}{c|}{\begin{tabular}[c]{@{}c@{}}\textbf{62.1}\\ 62.2\end{tabular}}                                                   & \multicolumn{1}{c|}{\begin{tabular}[c]{@{}c@{}}43.1\\ 43.4\end{tabular}}                                                   \\ \hline
\begin{tabular}[c]{@{}c@{}}RF+WDR\\ Ensemble\end{tabular}       & \multicolumn{1}{c|}{\begin{tabular}[c]{@{}c@{}}26.0\\ \textbf{25.9}\end{tabular}} & \multicolumn{1}{c|}{\begin{tabular}[c]{@{}c@{}}31.5\\ \textbf{30.8}\end{tabular}} & \multicolumn{1}{c|}{\begin{tabular}[c]{@{}c@{}}62.2\\ \textbf{62.1}\end{tabular}} & \begin{tabular}[c]{@{}c@{}}\textbf{41.8}\\ 41.9\end{tabular} \\ \hline
\end{tabular}}
\label{table:clm_results}
\end{table}

\subsubsection{Primary Experimental Results}
Table~\ref{table:clm_results} presents the entire results of the primary experiments (6 models on 4 datasets). Results show that, with the exception of TT-based models on W2, our proposed $N$-gram CLMs consistently either match or surpass the baseline CLMs, even without the ensemble method. Remarkably, WDR $N$-gram CLMs generally improve performance on top of the simple $N$-gram CLMs. Upon applying our proposed ensemble method, they generally exhibit improvements over their non-ensemble counterparts, except the models trained on W103. Notably, the effect of ensemble method is relatively significant in smaller datasets (PTB and W2) in contrast to larger datasets (T8 and W103). Based on these results, we argue that our proposed methods have actual advantages on various models and datasets for the CLM task.

\begin{table}[]
\centering
\caption{Experiment results of NMTs on several benchmark datasets. We used translations of TED and TEDx talks for IWSLT14 En-De. Also, we used Newstest18 and Newstest14 for WMT18 En-Tr and WMT14 En-De, respectively. The left and right numbers of `/' mean En-to-\textit{(De or Tr)} and \textit{(De or Tr)}-to-En translation results, respectively.}
\resizebox{0.43\textwidth}{!}{\begin{tabular}{|c|ccc|}
\hline
\multirow{2}{*}{Model} & \multicolumn{3}{c|}{BLEU Scores}           \\ 
\cline{2-4} & \multicolumn{1}{c|}{IWSLT} & \multicolumn{1}{c|}{WMT14} & WMT18 \\ 
\hhline{|====|}  
TF & \multicolumn{1}{c|}{27.6/32.5}                                                              & \multicolumn{1}{c|}{26.5/30.4}                              & \textbf{11.9}/18.2                                       \\ \hline
BOW NMT                                                     & \multicolumn{1}{c|}{27.5/32.3}                                                                     & \multicolumn{1}{c|}{26.3/30.4}                                              & \textbf{11.9}/18.3                                              \\ \hline
\begin{tabular}[c]{@{}c@{}}TF+Sim\\ Ensemble\end{tabular} & \multicolumn{1}{c|}{\begin{tabular}[c]{@{}c@{}}28.0/33.0\\ \textbf{28.3}/33.4\end{tabular}}                        & \multicolumn{1}{c|}{\begin{tabular}[c]{@{}c@{}}26.2/30.9\\ 26.3/31.0\end{tabular}} & \begin{tabular}[c]{@{}c@{}}11.6/18.2\\ 11.6/18.3\end{tabular} \\ \hline
\begin{tabular}[c]{@{}c@{}}TF+WDR\\ Ensemble\end{tabular} & \multicolumn{1}{c|}{\begin{tabular}[c]{@{}c@{}}27.9/33.5\\ \textbf{28.3}/\textbf{34.0}\end{tabular}} & \multicolumn{1}{c|}{\begin{tabular}[c]{@{}c@{}}\textbf{26.7}/31.1\\ \textbf{26.7}/\textbf{31.2}\end{tabular}} & \begin{tabular}[c]{@{}c@{}}11.8/18.5\\ \textbf{11.9}/\textbf{18.8}\end{tabular} \\ \hline
\end{tabular}}
\label{table:nmt_results}
\end{table}

\subsection{Neural Machine Translation}
\label{subsec:nmt_experimental_results}
\subsubsection{Data Description}
Since NMT includes language modeling as a part of the decoder, we view the NMT could be an appropriate additional experimental task to demonstrate the effectiveness of our proposed approach in addition to the main CLM tasks.  We conducted NMT experiments on several datasets: `IWSLT14 English-German'(En-De, 160K training pairs) \citep{hwang2023integrating}, `WMT14 English-German'(En-De, 3.9M training pairs) \citep{vaswani2017attention}, and `WMT18 English-Turkish' (En-Tr, 207K training pairs) \citep{bojar2018findings}. We used the same preprocessing, tokenization and subword byte-pair encoding methods with \citep{ott2019fairseq}. We used 10K, 10K, 32K most frequents subwords to organize vocabularies for datasets, respectively.

\subsubsection{Models and Training}
\label{subsubsec:nmt_models}
As a baseline, we used our implementation of Transformer (TF) \citep{vaswani2017attention} in the encoder-decoder architecture. We used the small Transformer for the `IWSLT14 En-De' and `WMT18 En-Tr' datasets,
and the base Transformer for the `WMT14 En-De' dataset. The total number of parameters of small and base TF baselines are 32M and 77M, respectively. We applied our simple and WDR $N$-gram CLM methods onto the decoder parts of the baselines, `TF+Sim' and `TF+WDR'. Each additional MLP layer in our simple and WDR methods increases the number of parameters by around 0.5M. Information about the models and how TF models are optimized can be found in the columns labeled 'Small Enc-Dec TF NMT' and 'Base Enc-Dec TF NMT' in Table ~\ref{table:setting_for_transformers}. Also, the hyperparameters ($N$ and $\lambda$) for `TF+Sim' and `TF+WDR' are described in the `NMT Task' column of Table~\ref{table:configurations_of_our_approach} (in Appendix \ref{subsec:experimental_settings}).


As a more closely related baseline, bag-of-words (BOW) NMT was proposed to predict the whole words in the context of the original NMT task \citep{ma2018bag}. However, their approach was not applied to the TF architecture, and they evaluated the model only on the English-Chinese translation dataset of NIST. To ensure a fair comparison, we re-implemented BOW NMT based on our TF architecture and compared with our proposed method. Following their prescribed approach, we integrated the computed loss of whole words prediction into the original loss.

\subsubsection{BLEU Results}
Table~\ref{table:nmt_results} presents the experiment results of the models on each testset with SacreBLEU \citep{post2018call} as the evaluation metric. Our proposed `TF+Sim' and `TF+WDR' models exhibit usually enhanced performances compared to the `TF' and `BOW NMT' baselines. `TF+WDR' always outperforms its counterpart of `TF+Sim'. Notably, the integration of the ensemble method from both of `TF+Sim' and `TF+WDR' further increases performances. Specifically, we note that `TF+WDR' with ensemble method improved performances by 0.7~1.5 BLEU scores compared to `TF' baseline on the both translation directions of `IWSLT14 En-De', and German-to-English translations of `WMT14 En-De' testsets. 

To explain why $N$-gram prediction approaches are more effective for German-to-English translation compared to English-to-German translation in `IWSLT14 En-De' and `WMT14 En-De' experiments, we hypothesize that the difference in word diversity between the two languages plays a role. We analyzed the `WMT14 En-De' training dataset (subword-level tokenized) and found that English has around 33.6K unique unigrams and 6.7M unique bigrams, while German has around 34.9K unique unigrams and 9.3M unique bigrams. This suggests that German-to-English translation might have simpler local dependencies to learn compared to English-to-German translation due to the lower number of unique bigrams. Considering simple local dependencies might lead to the over-fitting problem, we believe that this is a potential reason why $N$-gram prediction approaches, which can help mitigate over-fitting to local dependencies, are more effective for German-to-English translation.

\section{Conclusion}
In this work, we have constructed an advanced $N$-gram prediction framework tailored specifically to causal language modeling. 
In addition to the construction of this framework, our work includes the introduction of new strategies for providing diverse target representations and an ensemble method over the predicted $N$ words. Extensive experiments on language modeling and neural machine translation have confirmed the practical benefits of the proposed method.

\section{Limitations}
Given the demonstrated performance improvements of the WDR-based $N$-gram CLM, we tried to apply the WDR method to other tasks beyond CLM, such as the MLM task. In addition to the standard loss function of MLM, which involves predicting the masked word \citep{devlin2018bert}, we added new loss terms to predict $n$-level WDR target representations of the masked position. For this experiment, we utilized the CrammedBERT model \citep{geiping2023cramming}, a streamlined variant of BERT that facilitates faster pre-training while maintaining competitive performance on the GLUE benchmark. We integrated the WDR approach into this model and conducted a comparative analysis with the original CrammedBERT configuration. Further experimental details are provided in Appendix \ref{subsec:mlm_experiment}.

Table~\ref{table:mlm_results} (in Appendix \ref{subsec:mlm_experiment}) presents the results of our experiments comparing CrammedBERT and the applications of WDR models on the GLUE test set. While the application of 2-level WDR resulted in a 1.0 point increase in the average GLUE score, the performance benefits of the WDR method is less consistent across individual sub-tasks compared to the benefits observed in the CLM tasks. We attribute this result to the fundamental difference between the CLM and MLM tasks. Specifically, in MLM, when the WDR method combines the masked word embedding with the embeddings of the next words, such information is already provided as input. This partial visibility of the target representation might lead to an unexpected optimization behavior, such as the model disproportionately focusing on the right-side (future) context which is incorporated in the target, rather than considering the entire context.

Since there are prior works for $N$-gram prediction within the MLM framework \citep{sun2019ernie, joshi2020spanbert, xiao2020ernie, qi2020prophetnet}, we believe we can apply the WDR method to the prior works by combining the only masked words when WDR is calculated to solve the aforementioned issue. We expect that the high gradient diversity characteristic of the WDR method may offer additional benefits to the prior MLM framework.

\section*{Acknowledgements}

\bibliography{reference}

\onecolumn
\appendix

\section{Appendix}
\label{sec:appendix}

\subsection{Proof of Eq.\eqref{eq:n_level_WDR_w_orgemb}}
\label{subsec:proof_n_level_wdr_w_orgemb}
We provide a proof of Eq.\eqref{eq:n_level_WDR_w_orgemb} with the induction method. To avoid confusion, we temporarily change the notation of $\Delta_n\mathbf{x}^e_t$ in conjecture Eq.\eqref{eq:n_level_WDR_w_orgemb} to $\hat{\Delta}_n\mathbf{x}^e_t$ until it is proved. Based on the definitions of the 1 and $n$-level WDR, Eq.\eqref{eq:1level_wdr} and Eq.\eqref{eq:n_level_wdr}, we can verify the initial condition, that is $n=1$, holds as follows:
\begin{align}
    \Delta_{1}\mathbf{x}^e_t&=\mathbf{x}^e_{t+1}-\mathbf{x}^e_{t} \nonumber \\
    &=\binom{1}{0}(-1)^0\mathbf{x}^e_{t+1}+\binom{1}{1}(-1)^1\mathbf{x}^e_{t} \nonumber \\
    &= \sum_{i=0}^{1}\binom{1}{i}(-1)^i\mathbf{x}^e_{t+(1-i)} \nonumber \\
    &=\hat{\Delta}_{1}\mathbf{x}^e_t. \nonumber
\end{align}
Therefore, the conjecture holds for the initial condition. Then, by following the induction method, we assume the conjecture at $n$-level is true, that is $\hat{\Delta}_{n}\mathbf{x}^e_t=\Delta_{n}\mathbf{x}^e_t$. Then, the $(n+1)$-level WDR from the definition Eq.\eqref{eq:n_level_wdr} is derived to $\Delta_{n+1}\mathbf{x}^e_t=\Delta_{n}\mathbf{x}^e_{t+1}-\Delta_{n}\mathbf{x}^e_{t}=\hat{\Delta}_{n}\mathbf{x}^e_{t+1}-\hat{\Delta}_{n}\mathbf{x}^e_{t}$. Each term is derived as follows:
\begin{align}
    \hat{\Delta}_{n}\mathbf{x}^e_{t+1}&=\binom{n}{0}(-1)^0\mathbf{x}^e_{t+n+1}+\binom{n}{1}(-1)^1\mathbf{x}^e_{t+n}+ \nonumber \\
    &  \cdots +\binom{n}{n-1}(-1)^{n-1}\mathbf{x}^e_{t+2}+\binom{n}{n}(-1)^{n}\mathbf{x}^e_{t+1}, \nonumber \\
    -\hat{\Delta}_{n}\mathbf{x}^e_{t}&=\binom{n}{0}(-1)^1\mathbf{x}^e_{t+n}+\binom{n}{1}(-1)^2\mathbf{x}^e_{t+n-1}+ \nonumber \\
    &  \cdots +\binom{n}{n-1}(-1)^{n}\mathbf{x}^e_{t+1}+\binom{n}{n}(-1)^{n+1}\mathbf{x}^e_{t}, \nonumber \\
    \hat{\Delta}_{n}\mathbf{x}^e_{t+1}-\hat{\Delta}_{n}\mathbf{x}^e_{t}&=\binom{n}{0}(-1)^0\mathbf{x}^e_{t+n+1}+\left(\binom{n}{0}+\binom{n}{1}\right)(-1)^1\mathbf{x}^e_{t+n}+ \nonumber \\
    &  \cdots+\left(\binom{n}{n-1}+\binom{n}{n}\right)(-1)^n\mathbf{x}^e_{t+1}+\binom{n}{n}(-1)^{n+1}\mathbf{x}^e_{t} \nonumber \\
    &=\binom{n+1}{0}(-1)^0\mathbf{x}^e_{t+n+1}+\binom{n+1}{1}(-1)^1\mathbf{x}^e_{t+n}+ \nonumber \\
    &  \cdots+\binom{n+1}{n}(-1)^n\mathbf{x}^e_{t+1}+\binom{n+1}{n+1}(-1)^{n+1}\mathbf{x}^e_{t} \nonumber \\
    &=\sum^{n+1}_{i=0}\binom{n+1}{i}(-1)^i\mathbf{x}^e_{t+(n+1-i)} \nonumber \\
    &=\hat{\Delta}_{n+1}\mathbf{x}^e_{t}. \nonumber
\end{align}
Note that the binomial coefficient, $\binom{n}{i}$, is the $n$-th row and $i$-th value of Pascal's triangle, and it satisfies $\binom{n}{i-1}+\binom{n}{i}=\binom{n+1}{i}$. Based on this outcome, the conjecture holds for $(n+1)$-level if the $n$-level is true. Therefore, the conjecture is proved.

\subsection{Experiment Details}
\label{subsec:experimental_settings}

\begin{table}[]
\caption{Model and optimizer configurations of Transformer architectures used in the preliminary experiment of CLM and NMT tasks. We used the same notation for model configurations as in \citep{vaswani2017attention}, except the number of layers (\# of Layers) and multi-head attention's heads (\# of Heads). `ISRS' means the inverse square root learning rate scheduler \citep{ott2019fairseq} and `\# of Tokens' indicates the total number of tokens in a mini-batch at each iteration.}
\centering
\resizebox{0.45\textwidth}{!}{\begin{tabular}{|c|c|c|c|}
\hline
Config.         & \begin{tabular}[c]{@{}c@{}}Small\\ Enc.\\ TF CLM\end{tabular} & \begin{tabular}[c]{@{}c@{}}Small \\ Enc-Dec\\ TF NMT\end{tabular} & \begin{tabular}[c]{@{}c@{}}Base\\ Enc-Dec \\ TF NMT\end{tabular} \\ \hhline{|====|}
$d_{model}$    & 256     & 512          & 512         \\ \hline
$d_{ff}$       & 2100    & 1024         & 2048        \\ \hline
$d_k=d_v$       & 64      & 64           & 64          \\ \hline
$P_{drop}$     & 0.3     & 0.3          & 0.1         \\ \hline
$\epsilon_{ls}$ & 0.1     & 0.1          & 0.1         \\ \hline
\# of Layers    & 6       & 6            & 6           \\ \hline
\# of Head      & 4       & 4            & 8           \\ \hline
Optimizer       & Adam    & Adam         & Adam        \\ \hline
Learning Rate   & 0.00025 & 0.0005       & 0.001       \\ \hline
Scheduler       & None    & ISRS         & ISRS        \\ \hline
\# of Tokens    & 4K      & 4K           & 25K         \\ \hline
Patience        & 50      & 50           & 50          \\ \hline
\end{tabular}}
\label{table:setting_for_transformers}
\end{table}

\begin{table}[]
\caption{Changed configurations from the original Tensorized Transformer and Reformer \citep{ma2019tensorized, kitaev2020reformer}. We note that `\# of Tokens' indicates the total number of tokens in a mini-batch at each iteration.}
\centering
\resizebox{0.8\textwidth}{!}{\begin{tabular}{|c|ccc|cc|}
\hline
\multirow{2}{*}{Dataset} & \multicolumn{3}{c|}{Tensorized Transformer}                                                     & \multicolumn{2}{c|}{Reformer}                               \\ \cline{2-6} 
                         & \multicolumn{1}{c|}{\# of Tokens} & \multicolumn{1}{c|}{\# of Layers} & Learning Rate           & \multicolumn{1}{c|}{\# of Tokens} & Learning Rate           \\ \hhline{|======|}
PTB                      & \multicolumn{1}{c|}{3,840}           & \multicolumn{1}{c|}{3}            & \multirow{4}{*}{0.0025} & \multicolumn{1}{c|}{16,384}       & \multirow{4}{*}{0.0001} \\ \cline{1-3} \cline{5-5}
WikiText-2               & \multicolumn{1}{c|}{3,840}           & \multicolumn{1}{c|}{3}            &                         & \multicolumn{1}{c|}{8,192}        &                         \\ \cline{1-3} \cline{5-5}
Text8                    & \multicolumn{1}{c|}{4,800}           & \multicolumn{1}{c|}{6}            &                         & \multicolumn{1}{c|}{512}          &                         \\ \cline{1-3} \cline{5-5}
WikiText-103             & \multicolumn{1}{c|}{4,800}           & \multicolumn{1}{c|}{6}            &                         & \multicolumn{1}{c|}{512}          &                         \\ \hline
\end{tabular}}
\label{table:changed_settings_for_primary_baselines}
\end{table}

\begin{table*}[]
\caption{Configurations of our proposed $N$-gram approaches: $N$ and $\lambda$, used in the primary experiments of the CLM task and experiments of the NMT task. }
\centering
\resizebox{0.9\textwidth}{!} { \begin{tabular}{|cccccc|ccccc|}
\hline
\multicolumn{6}{|c|}{CLM Task}                                                                                                                                                              & \multicolumn{5}{c|}{NMT Task}                                                                                                       \\ \hline
\multicolumn{1}{|c|}{\multirow{2}{*}{Model}} & \multicolumn{1}{c|}{\multirow{2}{*}{Config.}} & \multicolumn{4}{c|}{Dataset}                                                                 & \multicolumn{1}{c|}{\multirow{2}{*}{Model}}  & \multicolumn{1}{c|}{\multirow{2}{*}{Config.}} & \multicolumn{3}{c|}{Dataset}         \\ \cline{3-6} \cline{9-11} 
\multicolumn{1}{|c|}{}                       & \multicolumn{1}{c|}{}                         & \multicolumn{1}{c|}{PTB}   & \multicolumn{1}{c|}{W2}    & \multicolumn{1}{c|}{T8}    & W103  & \multicolumn{1}{c|}{}                        & \multicolumn{1}{c|}{}                         & \multicolumn{1}{c|}{IWSLT14} & \multicolumn{1}{c|}{WMT14} & WMT18 \\ 
\hhline{|======|=====|}
\multicolumn{1}{|c|}{TT+Sim}                 & \multicolumn{1}{c|}{N/$\lambda$}                 & \multicolumn{1}{c|}{2/0.2} & \multicolumn{1}{c|}{4/0.2} & \multicolumn{1}{c|}{3/0.2} & 2/0.1 & \multicolumn{1}{c|}{\multirow{2}{*}{TF+Sim}} & \multicolumn{1}{c|}{N}                        & \multicolumn{1}{c|}{3} & \multicolumn{1}{c|}{2}      & 2     \\ \cline{1-6} \cline{8-11} 
\multicolumn{1}{|c|}{TT+WDR}                 & \multicolumn{1}{c|}{N/$\lambda$}                 & \multicolumn{1}{c|}{2/0.4} & \multicolumn{1}{c|}{4/0.3} & \multicolumn{1}{c|}{3/0.1} & 2/0.1 & \multicolumn{1}{c|}{}                        & \multicolumn{1}{c|}{$\lambda$}                   & \multicolumn{1}{c|}{0.3}     & \multicolumn{1}{c|}{0.1} & 0.2   \\ \hline
\multicolumn{1}{|c|}{RF+Sim}                 & \multicolumn{1}{c|}{N/$\lambda$}                 & \multicolumn{1}{c|}{4/0.2} & \multicolumn{1}{c|}{2/0.2} & \multicolumn{1}{c|}{3/0.1} & 4/0.1 & \multicolumn{1}{c|}{\multirow{2}{*}{TF+WDR}} & \multicolumn{1}{c|}{N}                        & \multicolumn{1}{c|}{3} & \multicolumn{1}{c|}{2}       & 2     \\ \cline{1-6} \cline{8-11} 
\multicolumn{1}{|c|}{RF+WDR}                 & \multicolumn{1}{c|}{N/$\lambda$}                 & \multicolumn{1}{c|}{4/0.1} & \multicolumn{1}{c|}{2/0.3} & \multicolumn{1}{c|}{3/0.1} & 4/0.1 & \multicolumn{1}{c|}{}                        & \multicolumn{1}{c|}{$\lambda$}                   & \multicolumn{1}{c|}{0.5} & \multicolumn{1}{c|}{0.1}    & 0.3   \\ \hline
\end{tabular}}
\label{table:configurations_of_our_approach}
\end{table*}

We trained the models described in Sec. \ref{subsubsec:clm_models} and Sec. \ref{subsubsec:nmt_models} following the configurations described in Table~\ref{table:setting_for_transformers} for Transformer-based models, `TF', and the configurations reported in the previous works' papers \citep{ma2019tensorized, kitaev2020reformer} with several changes as described in Table ~\ref{table:changed_settings_for_primary_baselines} for the primary CLM baselines, `TT' and `RF'. For Transformer-based models' experiments, we saved the best checkpoint based on the validation results. We early stopped the training whenever the model does not beat its previous best performance for the `Patience' times on the validation \citep{heo2023shared}. For the primary CLM baselines, we followed the pre-defined total training iterations. Table~\ref{table:configurations_of_our_approach} describes the specific configurations, such as $N$ and $\lambda$, we used for our proposed $N$-gram CLMs, simple-based and WDR-based. 

About the information of our computational environment, we used a single NVIDIA RTX3090 GPU for the large CLM datasets, such as T8 and W103, and a GTX1080Ti GPU for the small CLM datasets, such as PTB and W2. On average, they took 1 day and 3 hours, respectively, for training. We used 4x NVIDIA RTX3090 GPUs for the large NMT datasets, such as WMT14 English-German, and 2x GTX1080Ti GPUs for the small NMT datasets, such as IWSLT14 English-German and WMT18 English-Turkish. On average, they took 3 days for training.

\begin{table*}[t]
\centering
\caption{Experiment results of MLMs on the GLUE task. We used the same metrics with \citep{geiping2023cramming} for each sub-task in GLUE.}
\resizebox{0.9\textwidth}{!}{\begin{tabular}{|c|c|c|c|c|c|c|c|c|c|}
\hline
Model       & MNLI               & SST-2         & STSB          & RTE           & QNLI          & QQP           & MRPC          & CoLA          & GLUE Avg.     \\ 
\hhline{|==========|}
CrammedBERT & 78.5/79.0          & \textbf{90.0} & 82.3          & \textbf{57.4} & 85.7          & 85.7          & 85.2          & 28.5          & 74.6          \\ \hline
+1-level WDR      & 78.3/\textbf{79.2} & 88.2          & 80.0          & 54.2          & 85.9          & 85.7          & 84.4          & 30.3          & 74.0          \\ \hline
+2-level WDR    & 78.6/79.1          & 88.4          & \textbf{82.4} & 55.2          & 85.5          & 85.8          & \textbf{86.9} & \textbf{38.6} & \textbf{75.6} \\ \hline
+3-level WDR      & \textbf{78.8}/79.1 & 89.0          & 81.8          & 56.3          & \textbf{86.4} & \textbf{85.9} & 85.6          & 32.8          & 75.1          \\ \hline
\end{tabular}}
\label{table:mlm_results}
\end{table*}

\subsection{Masked Language Modeling Experiment}
\label{subsec:mlm_experiment}
We adhered to the environmental settings established by CrammedBERT \citep{geiping2023cramming} for all aspects of our study, including dataset preprocessing, model configurations, pre-training, fine-tuning procedures, and evaluations. Comprehensive details of these settings can be found in the associated GitHub repository\footnote{https://github.com/JonasGeiping/cramming}. Building on the CrammedBERT architecture, we apply the WDR method that is analogous to the method conducted in our WDR-based $N$-gram CLM experiment. Specifically, we utilized $N$ additional MLP layers designed to predict $n$-level WDRs alongside the original word embedding at the masked position. These $n$-level WDRs are calculated by composing the next words of the masked word. The final loss is computed as the average of the original loss and the additional losses derived from the WDR method, with the original and additional losses being averaged unequally, as described in Section \ref{subsec:simple_n_gram_clm}.

Table~\ref{table:mlm_results} presents the experimental results for CrammedBERT and our proposed models, evaluated on the GLUE test set following fine-tuning. We varied the number of grams, $N$, from 1 to 3. The results indicate that the application of 2-level WDR yields an increase of 1.0 point in the average GLUE score. However, the performance improvements across individual sub-tasks are not consistently superior; in some cases, they were similar to or worse than the baseline.

\end{document}